\documentclass[10pt, a4paper]{article}
\usepackage{lrec}
\usepackage{multibib}
\newcites{languageresource}{Language Resources}

\usepackage{booktabs}
\usepackage{covington} 
\usepackage{graphicx}
\usepackage{epstopdf}
\usepackage{amssymb}
\usepackage{todonotes}

\usepackage[utf8]{inputenc}

\usepackage{tabularx}

\usepackage{soul}
\usepackage{url} 
\usepackage{xstring}
\usepackage{enumitem}

\newcommand{\SDInlineExample}[1]{{\small\textit{\textsf{``#1''}}}}

\title{A Corpus Study and Annotation Schema for Named Entity Recognition and Relation Extraction of Business Products}

\name{Saskia Sch\"{o}n, Veselina Mironova, Aleksandra Gabryszak, Leonhard Hennig}

\address{DFKI GmbH \\
         Berlin, Germany\\
         \{firstname.lastname\}@dfki.de\\}

\abstract{Recognizing non-standard entity types and relations, such as B2B products, product classes and their producers, in news and forum texts is important in application areas such as supply chain monitoring and market research. However, there is a decided lack of annotated corpora and annotation guidelines in this domain. In this work, we present a corpus study, an annotation schema and associated guidelines, for the annotation of product entity and company-product relation mentions. We find that although product mentions are often realized as noun phrases, defining their exact extent is difficult due to high boundary ambiguity and the broad syntactic and semantic variety of their surface realizations. We also describe our ongoing annotation effort, and present a preliminary corpus of English web and social media documents annotated according to the proposed guidelines. 
\\ \newline 
}

\begin{document}

\maketitleabstract

\section{Introduction}
\label{sec:intro}
Recognizing non-standard entity and relation types is an important task in many real-world information extraction applications like relation extraction, knowledge base construction and question answering. In areas such as market research and supply chain management, many companies would benefit from systems that automatically and continuously acquire up-to-date information about producers, vendors and other suppliers of specific parts, products, new technologies and components. Similarly, the construction of knowledge graphs that store supplier and vendor relationships would clearly benefit from information extraction approaches by reducing the manual effort required to create and maintain such databases. For example, in both scenarios it would be useful to extract information about e.g.\ a \emph{CompanyProvidesProduct} relation from a news text like  \SDInlineExample{Sensata Technologies' products include speed sensors, motor protectors, and magnetic-hydraulic circuit breakers}, where the \emph{product} argument refers to a non-consumer product or product class entity such as \SDInlineExample{speed sensors} or \SDInlineExample{magnetic-hydraulic circuit breakers}. 

However, when it comes to such specific domains, developing named entity recognition algorithms is severely hampered by the lack of publicly available training data and the difficulty of accessing existing dictionary-type resources, such as product catalogs. Many available named entity recognition corpora consist of general news articles~\cite{tjong2003,doddington2004,weischedel2013}, while information about B2B products is typically available on non-journalistic, specialized web portals and forums. Product mentions, as in the example above, are often general noun phrases, instead of proper names, which increases the difficulty of detecting them using gazetteer-based approaches. In addition, relational information about companies and their products is very limited in freely available knowledge bases (KB), such as Freebase~\cite{bollacker2008}, Wikidata~\cite{vrandecic2014}, or DBpedia~\cite{auer2007}, since these KBs are in large parts based on Wikipedia, which aims to exclude commercial, non-encyclopedic information. For example, DBpedia contains only approximately $60,000$ triples for the \emph{CompanyProvidesProduct} relation.

To address these problems, and to gain a better understanding of product mentions and their linguistic properties, in this study we first collect a large number of noisy product mentions. This is achieved with a bootstrapping approach that uses a set of manually defined lexical patterns for the relation \emph{CompanyProvidesProduct} (Section~\ref{sec:bootstrap}). We analyze the resulting set of mentions, and find that they often include extraneous linguistic material that should not be considered a part of the product extent, such as prepositional phrases and appositions. Consequently, we develop an annotation schema for \emph{product} mentions and the \emph{CompanyProvidesProduct} relation, in order to guide the manual annotation of texts (Section~\ref{sec:schema}). We are currently building a corpus of English web and social media documents with annotations for \emph{product} entity and \emph{CompanyProvidesProduct} relation mentions based on these annotation guidelines. We will make a first version of the corpus available to the community (Section~\ref{sec:corpus}). The overall goal of our work is to make extraction of non-standard, B2B products and relations from unstructured text easier and more reliable.

%
%
%
%
%
%
%
%

\section{Related Work}
\label{sec:relwork}
Most research in Named Entity Recognition (NER) has focused on common entity types, such as persons, organizations, and locations~\cite{tjong2003,finkel2005,derczynski2016}, and numeric types like date and time expressions~\cite{stroetgen2013}. Only a few corpora cover other entity types, such as geopolitical entities and facilities~\cite{doddington2004,weischedel2013}. Corpora that include product annotations are rare: the BBN corpus covers (consumer) products mentioned in Wall Street Journal news articles~\cite{weischedel2005}. Liu et al.~\shortcite{liu2011} describe a corpus of tweets that has been annotated with products, but the dataset is not publicly available. Recent research in fine-grained NER has produced distantly~\cite{ling2012,weischedel2013} or weakly supervised datasets~\cite{ni2017} using Freebase and Wikipedia, which therefore inherit the  coverage  and specificity limitations of these resources. The datasets include products and their subtypes, but the entities are generally consumer products, such as cars, mobile phones, and games. In the case of Ni et al.~\shortcite{ni2017}, the dataset is not publicly available. 

For relation extraction, there exist only very few datasets that have been manually annotated with company-product-related information. FrameNet~\cite{baker1998} contains example sentences marked up with frames that provide information that is similar to the \emph{CompanyProvidesProduct} relation, such as the \emph{Business}, \emph{Commerce\_sell} and \emph{Manufacturing} frames. The SemEval 2010 Task 8 contains 968 sentences annotated with pairs of nominals for mentions of the \emph{Product-Producer} relation~\cite{hendrickx2010}. While the relation's name suggests similarity to our dataset, the scope in the SemEval dataset is much broader and includes any kind of production, e.g.\ blisters caused by a herpes virus, children ``produced'' by their parents, or questions asked by journalists.

The ACE guidelines for English relations~\cite{ldc2005} describe the relation type \emph{Agent-Artifact}, which ``applies when an agent owns an artifact, has possession of an artifact, uses an artifact, or caused an artifact to come into being.'' However, the non-organization argument of the corresponding relation definition only allows \emph{facility}-type entities, and does not mention products. 

\section{Bootstrapping Product Annotation}
\label{sec:bootstrap}
\begin{table*}[ht!]
\footnotesize
\begin{tabular}{lp{7.5cm}}
\toprule
  \textbf{Pattern} & \textbf{Example}  \\ 
\midrule
ORG\emph{'s} PRO:\{$<$VBG\textbar NN.*\textbar JJ\textbar CD$>$*$<$NN.*$>$+$<$NN.*\textbar JJ\textbar CD$>$*\} & \underline{BMW}'s [1-Series Convertible] is a stylish convertible. \\
PRO \emph{by} ORG & [Intuition Executive] by \underline{Honeywell} collects and analyzes large amounts of data. \\
ORG [\emph{to produce\textbar to manufacture\textbar to develop\textbar\ldots}] PRO & \underline{Sensata Technologies} develops [sensors] and [controls]. \\
ORG [\emph{to be}] [\emph{producer of\textbar maker of\textbar\ldots}] PRO & \underline{Amazon} is a vendor of [books] and [technology products]. \\
ORG [\emph{to be}] [\emph{a\textbar the\textbar an}] PRO [\emph{producer\textbar provider\textbar supplier\textbar ...}] & \underline{Apple} and \underline{Samsung} are [smartphone] providers.\\
\bottomrule
\end{tabular}
\caption{Example bootstrap patterns for the relation \emph{CompanyProvidesProduct} used for pre-annotating product mentions. Company arguments are underlined, product arguments enclosed in brackets. For brevity, the chunking pattern applied to extract potential product mentions is only shown in the first row of the table.}
\label{tab:patterns}
\end{table*}

This section presents the data sources used in this work, and the pattern-based bootstrapping approach used for pre-annotating products. 

\subsection{Source Datasets}
\begin{figure}[t!]
     \centering
     \includegraphics[width=\columnwidth,trim={0 0 0 0},clip]{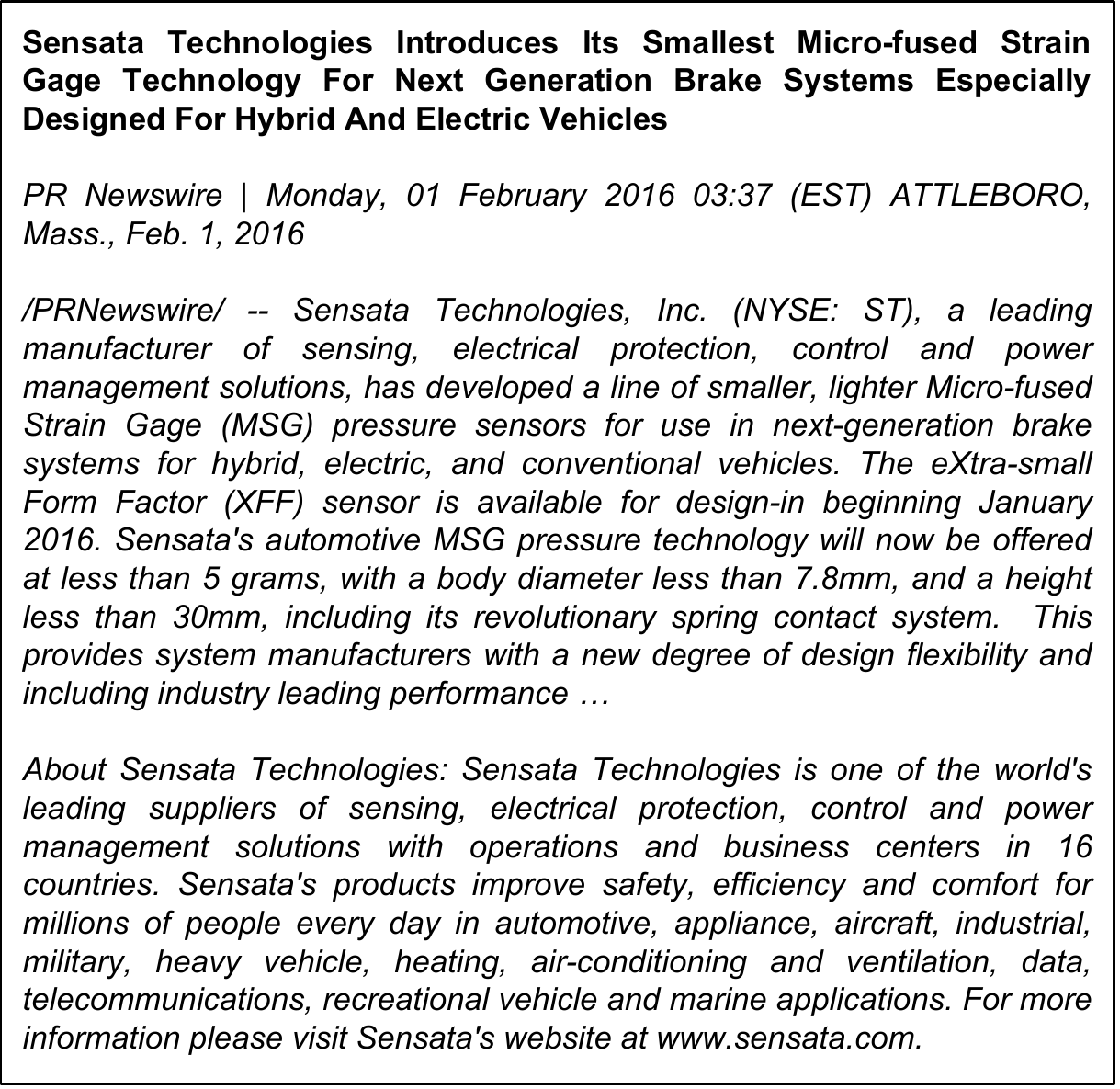}
     \caption{Example document relating B2B product and supplier information.}
     \label{fig:example_news}
 \end{figure}

We collected a large dataset of web pages from business news portals, company home pages, and special interest forums, as well as posts from social media channels such as Twitter and Facebook. Web pages and forums were crawled based on an iteratively refined set of source URLs and keywords, such as company name lists. Similar keyword sets were also used for retrieving public Twitter and Facebook posts using the respective APIs of these services. The dataset was crawled over a period of 1.5 years, between July 2016 and December 2017, and consists of approximately $5.8$ million documents. We focused on B2B information related sites, but did not explicitly exclude documents reporting on consumer products. Figure~\ref{fig:example_news} shows an example document from the corpus. 

As can be expected, documents exhibit a large degree of linguistic variance, ranging from journalistic writing to colloquially formulated tweets. In addition, HTML-to-text conversion and boilerplate removal are far from perfect, resulting in extraneous and not-well-formed content. We noticed that B2B news sites often appended ``canned'' company summaries at the end of news articles, which typically contain a lot of useful information about the company's products. We included an example of this feature as the second paragraph of Figure~\ref{fig:example_news}.





\subsection{Pattern-based Product Pre-Annotation}

To gain a better understanding of the linguistic properties of the different types of product mentions in our dataset, we included an automatic pre-annotation step in our corpus analysis. Pre-annotation can help to decrease the duration of manual annotation and generally ease the annotation process~\cite{kwon2014}. We first developed a set of lexical patterns for the relation \emph{CompanyProvidesProduct}. We chose this approach since we are also interested in identifying instances of this relation, and since it significantly increases the precision of product mention identification (at the cost of recall). 

The \emph{CompanyProvidesProduct} relation maps a company (organization) to products created, manufactured, provided, distributed or vendored by this company (Section~\ref{sec:relation}). 
Table~\ref{tab:patterns} lists some example patterns.  Many of these patterns can be varied by exchanging the verb or verb nominalization used, e.g.\ \emph{produce, create, develop, make, manufacture, offer}. In total, we defined $13$ base patterns, yielding a total of $173$ surface patterns. We used a chunk parser to label potential product mentions. It matches noun phrases, optionally preceded or followed by adjectives or cardinal numbers, e.g.\ \SDInlineExample{high-resolution waveform analysis}, \SDInlineExample{High-Frequency 600mA DC-DC Buck Converter}, and \SDInlineExample{1500 ECL-PTU-208}. We included the \emph{VBG} tag as in rare cases, gerund verb forms may be part of the product mention, e.g.\ \SDInlineExample{communicating sensors}.

For pre-annotation, we randomly selected a set of $1,200$ documents from the source dataset. Each document's text was tokenized and part-of-speech tagged. Organizations were detected using Stanford NER~\cite{manning2014}. We then applied the patterns, which resulted in a total of $1,308$ potential product mention matches. 

While many of these matches did at least cover a product mention, many results were unsatisfactory because they included excess lexical material, as shown in the following examples (square brackets denote the correct mention extent):

\begin{example}
\small
	\begin{itemize}
		\item[a.] \textit{\textsf{*highly accurate [3D magnetic sensor]}}
		\item[b.] \textit{\textsf{*advanced [magnetic-hydraulic circuit breakers]}}
        \item[c.] \textit{\textsf{*Rambus' [R+] industry-standard [interface solutions]}}
	\end{itemize}
	\label{ex:1}
\end{example}

We also observed that even when syntactical extraction worked correctly in a way that only phrases containing a product mention were retrieved, there were differences in semantic quality that should be taken into account. An adjective in the same position, for example, can either be of no value to the categorial specification of the product, in which case it can be neglected, or it can be a crucial part of the category description:

\begin{example}
\small
	\begin{itemize}
 		\item[a.] \textit{\textsf{advanced [sensors]}}
		\item[b.] \textit{\textsf{[magnetic sensors]}}
    \end{itemize}
    \label{ex:2}
\end{example}

The extent to which these issues occurred seems to be closely linked with the specific product domain. These observations show that due to high boundary ambiguity as well as a broad syntactic and semantic variety of the surface variants of product mentions, it is necessary to define in advance which elements should be considered part of the extent of a product mention, and which should be excluded. 

%
%










\section{Annotation Guidelines} 
\label{sec:schema}

To formalize the annotation of \emph{product} mentions and \emph{CompanyProvidesProduct} relation mentions, we developed a set of annotations guidelines. For entity annotation, we base our guidelines on existing work, such as the ACE annotation guidelines~\cite{doddington2004} for labeling \emph{organizations} and \emph{companies}. We try to follow similar guidelines for the annotation of products, but transform and expand these as detailed below. 

Since many phrases in a typical document can be viewed as products or product classes (e.g.\ \SDInlineExample{mobile services}, \SDInlineExample{hotel chains}, \SDInlineExample{personal devices}), annotators limited their effort by adopting the following overall strategy for labeling a document: First, they annotated all name  mentions of \emph{organizations} and \emph{products}, as well as any coreferential nominal and pronominal mentions of these. Coreference information was added as an extra relation type \emph{Identity}. A single \emph{Identity} relation was created for each coreference chain, with a \emph{source} argument for the most precise name mention of an entity in a document, and \emph{target} arguments for all other mentions of this entity. In a second step, annotators searched for occurrences of \emph{CompanyProvidesProduct} relation mentions, and labeled (pro-) nominal product or product class references if they served as the argument of the identified relation mention. For example, annotators would label \SDInlineExample{sensors} and \SDInlineExample{controls} as \emph{product}s in the sentence \SDInlineExample{Sensata Technologies develops sensors and controls.} during this second step.\footnote{Only if these nouns were not labeled as coreferential to some product name mention in the first step, of course.} In a last step, coreferential mentions of these additional \emph{product} mentions were also labeled in the remainder of the document, even if they did not occur as an argument of a relation mention. The reasoning here is that an NER algorithm should encounter consistent labels for the same token sequence, e.g.\ if \SDInlineExample{smartphones} is labeled as a \emph{product} once, it should be labeled as a \emph{product} everywhere in the document. All other product or product class references, i.e.\ those that were not part of a \emph{CompanyProvidesProduct} relation mention, were not annotated as \emph{product} entity mentions. We chose this strategy to limit the annotation effort for the initial corpus. In addition, disagreement by the annotators was very high when annotating all noun phrases that could potentially be viewed as products.


\begin{table*}[ht!]
  \centering
  \small
\begin{tabular}{lp{3cm}p{10.8cm}} 
\toprule
           \textbf{Category} & \textbf{Example} & \textbf{Description} \\
           \midrule
          company name & [Dunlop] Sport M3 winters & This is the name of the company that provides the product when it is mentioned as part of the product name. It is usually found in the beginning of the product mention. It tends to be a proper noun that does not always follow orthographic rules and sometimes appears as an abbreviation (e.g. ``Tumblr'', ``BMW''). The complexity of a company name tends to be reduced to the main word(s) when it is part of the product mention, i.e., ``Toyota'' instead of ``Toyota Motor Corporation''. \\
          \midrule
          brand name & Apple [iPhone] 6S & The brand name is the name under which a certain product is marketed. Usually this is a proper name, however it is not always capitalized or otherwise orthographically correct. Like all categories that can be realized as a proper name, a brand name can include any word class, special characters and punctuation (e.g.\ ``FILL OR BUST!''). \\
         \midrule
        series &  VW Golf [VII] & This is the part of the product mention that denotes the series, generation, edition or model range. It is often realized as a number, sometimes as a name. Sometimes it includes the word ``series'' or ``generation'' or an equivalent abbreviation. Whether or not this can be part of a product mention is dependent on the domain.  \\
           \midrule
        model & BMW [i8] & This part of the product mention denotes the specific model of a product in a product series. It often consists of letters and numbers or a combination of the two.  \\
           \midrule
        trademark symbol &  McRib[\textsuperscript{\textregistered}] &  The trademark symbol usually appears right after the brand name.  \\
           \midrule
          type & Nike Air Max 2016 [running [shoes]] & The type of the product is the broader category or subcategory a product falls into. The type is usually a common noun and can often be found at the end of the product mention. The category can include an attribute that serves as a specifier to the noun and it is the category in which nonspecific terms such as ``product'' or ``solutions'' can be included, serving as the head of the product mention, but only if specified further (e.g. ``cosmetic product'').  \\
          \midrule
          feature & [2006] Ford Mustang [GT] Convertible [2-Door] \newline\newline
          Samsung Galaxy S7 [32 GB] [black]
          & Most other relevant aspects of products we will categorize as a feature. Since this is the vaguest category and the one that is most highly domain-dependent, it can be represented by a broad variety of linguistic manifestations. It includes elements such as the year of fabrication, colors, sizes, variants, and special features. Features can appear in almost any position in the product mention. \\
\bottomrule
\end{tabular}
\caption{Categories of product mention elements. Square brackets denote the extent of elements.}
\label{tab:annotation-categories}
\end{table*}


\subsection{Products and Product Mentions}
We define as a product any commercially available good, be it a finished product, a pre-product, or a part or component of a larger product. While the focus of this work is on non-consumer products, this definition also includes consumer products. A product does not have to be a tangible object, but can be a service or virtual object. 
Although they are semantically closely related, product-like entity mentions that refer to an industry sector or branch of business are not treated as products. The industry term can, however, be part of the product mention. Categories such as brand names and trademarks also often appear as part of the product mention. 

For the reliable extraction of a product mention its maximum extent must be pre-defined. This means that one has to identify both the elements included in the extent and those outside of the extent that often appear alongside the product mention. We will discuss the different elements a product mention can consist of, considering both semantic categories and their word class counterparts, as well as elements that are excluded from the product extent according to our annotation schema. 

The ways in which a product can be mentioned in a text are manifold: 
\begin{example}
\small
	\begin{itemize}
 		\item[a.] \textit{\textsf{vehicle}}
		\item[b.] \textit{\textsf{SUV}}
		\item[c.] \textit{\textsf{Land Cruiser}}
		\item[d.] \textit{\textsf{Toyota Land Cruiser}}
		\item[e.] \textit{\textsf{Toyota Land Cruiser 100 Series VX}}
		\item[f.] \textit{\textsf{Toyota Land Cruiser 100 Series VX SUV}}
    \end{itemize}
    \label{ex:3}
\end{example}

All of these examples are possible ways to refer to the same real-world product and could appear as the product argument in a relation expressing a product the company ``Toyota'' sells. \ref{ex:3}[a.] and \ref{ex:3}[b.] are rather vague, describing a product category, \ref{ex:3}[c.] and \ref{ex:3}[d.] are more specific, distinguishing the car from all other brands by all other companies, and \ref{ex:3}[e.] and \ref{ex:3}[f.] are so specific that the product cannot be confused with another model. 

Product mentions are generally realized as noun phrases, containing at least one proper noun or one common noun. As a proper noun, the head of the noun phrase can consist of individual letters or numbers or a series of numbers and/or letters:

\begin{example}
\small
	\begin{itemize}
 		\item[a.] \textit{\textsf{AP3405}}
		\item[b.] \textit{\textsf{1500 ECL-PTU-208}}
        \item[c.] \textit{\textsf{Samsung 14nm LPP Process}}
    \end{itemize}
    \label{ex:4}
\end{example}

Often, the noun is accompanied by further distinctive attributes that can appear in different word classes as illustrated in the following examples:

\begin{example}
\small
	\begin{itemize}
 		\item[a.] \textit{\textsf{smart sensors}} (adjective)
        \item[b.] \textit{\textsf{communicating sensors}} (verb, gerund)
        \item[c.] \textit{\textsf{vision sensors}} (common noun)
        \item[d.] \textit{\textsf{Hall sensors}}  (proper noun)
    \end{itemize}
    \label{ex:5}
\end{example}

\subsection{Elements of Product Mentions}

We found a limited set of elements that products usually consist of. This set can be subdivided into seven categories: \emph{company name, brand name, series, model, trademark symbol, type} and \emph{feature}. Not all of these elements appear in every product mention. Product mentions can vary strongly in length and complexity, from a single element (\ref{ex:6}[a.-c.]), to a combination of any of the categories (\ref{ex:6}[d.-f.]) to a coverage of all of the categories (\ref{ex:6}[g.]).

\begin{example}
\small
	\begin{itemize}
 		\item[a.] \textit{\textsf{sensors}} (type)
        \item[b.] \textit{\textsf{Kleenex}} (brand name)
		\item[c.] \textit{\textsf{Q7}} (model)
		\item[d.] \textit{\textsf{Audi Q7}} (company name and model)
		\item[e.] \textit{\textsf{Innocent Drinks smoothies}} (company name and type)
		\item[f.] \textit{\textsf{white iPhone 6}} (feature, brand name and model)
		\item[g.] \textit{\textsf{Toyota Land Cruiser 100 Series VX SUV diesel turbo}} (all of the above)
    \end{itemize}
    \label{ex:6}
\end{example}


As all of the examples used thus far have shown, some categories are more essential to a product mention than others. A product mention contains at least a common noun, representing the product \emph{type}, or a proper noun that can either refer to a \emph{brand name} or a specific \emph{model}.
Like brand names and models, the company name that often appears as part of the product mention (but is not essential to it) can also consist of a proper name and therefore include any kind of word class and even punctuation. While most of the seven categories can include or can be realized as nouns, \emph{series} and \emph{generations} as well as \emph{models} tend to consist of letters or numbers or combinations of the two. Table~\ref{tab:annotation-categories} lists the further specification of the individual categories as well as examples for each category. In the table, the different categories of elements that we consider part of product mentions are listed in the order in which they usually appear when a product mention contains more than one category, with the exception of the category \emph{feature} that can be found in any position.


Since it is not always apparent which category a part of a product mention falls into -- sometimes the same part of a product mention could be assigned to two or more categories -- the nested elements that constitute a product mention are currently not annotated, but only used by the annotator to determine the product mention's extent. 


\subsection{Elements Excluded from Product Mentions}

One of the major results of our analysis is that there are a number of elements that often appear alongside a product mention and may be mistaken as part of it. Unless they are included in the proper name of the product (usually the company name or the brand name part), articles, prepositional phrases and prepositions, relative clauses and appositions are never considered part of the product mention extent. A more detailed discussion of these elements can be found in our annotation guidelines. We will only go into detail here regarding the more interesting, less clear-cut cases, namely company names, adjectives and other attributive elements as well as conjunctions and punctuation elements. 

A company name that is used as the first argument in a \emph{CompanyProvidesProduct} relation mention is considered part of the product extent if it does not come with a possessive marker. 
Since the line between a company name and a brand name can be blurred, we follow this rule to differentiate between cases of a nested relation mention (a relation mention within the product mention) and separate mentions of company and product. Usually, punctuation between words marks a product mention's boundary. This is not the case for hyphens if they connect different elements of a product mention. Commas and linking conjunctions can also serve as connectors when they list different elements of the same product, such as features or attributes. They can, however, also list different products. This merges into the aspect of attributes that often precede the head of the noun phrase that is the product mention. As discussed before, adjectives and other attributive elements are not considered part of the product extent unless they serve to define the product further. If they do, but there is more than one attribute fulfilling that function, we have to differentiate between products that are described by several attributes on the one hand and different product mentions that share a head but are distinguished by the attributes on the other. In the former case, the commas or linking conjunctions are included in the extent of the product, in the latter case they are not, but two -- or more -- product mentions are annotated. The following examples illustrate this issue:

\begin{example}
\small
	\begin{itemize}
 		\item[a.] \textit{\textsf{[semiconductor] and [IP products]}}
        \item[b.] \textit{\textsf{[analog], [digital] and [mixed-signal integrated circuits]}}
		\item[c.] \textit{\textsf{[wireless and self-powered LED controls]}}
     \end{itemize}
    \label{ex:7}
\end{example}
Examples~\ref{ex:7}[a.] and [b.] contain attributes that are assigned to different products, whereas \ref{ex:7}[c.] illustrates the case of two different attributes that specify the same product.






\section{The CompanyProvidesProduct Relation}
\label{sec:relation}

The \emph{CompanyProvidesProduct} relation consists of two mandatory arguments, a \emph{company} (organization) and a \emph{product}, as well as of one optional argument, a \emph{trigger}.



A company can serve as the first argument if it is stated as the creator, manufacturer, provider, distributor or vendor of the product argument. The slot for the second mandatory argument can be filled by one or more product mentions (e.g.\ in the case of conjunctive enumerations). Trigger concepts are a generic class of annotations that cover lexical expressions (terms or phrases) or syntactical elements (e.g. possessive marker \textit{-s} or prepositional constructs) that indicate a specific event type. 

The annotators were instructed to annotate only relation instances mentioned within a sentence.  The following examples illustrate the relation annotation:

\begin{example}
\small
	\begin{itemize}
    \item [a.] \textit{\textsf{[Parkifi]\textsubscript{company} is a fast-growing technology company focused on [providing]\textsubscript{trigger} their customers with [real-time parking data]\textsubscript{product}}}
    \item [b.] \textit{\textsf{[Sensata Technologies Holding]\textsubscript{company} [produces]\textsubscript{trigger} [sensors]\textsubscript{product}}}
    \item [c.] \textit{\textsf{[BMW]\textsubscript{company}['s]\textsubscript{trigger} [Z3]\textsubscript{product}}}
    \item [d.] \textit{\textsf{[Intuition Executive]\textsubscript{product} [by]\textsubscript{trigger} [Honeywell]\textsubscript{company} collects and analyzes large amounts of data}}
    \item [e.] \textit{\textsf{[[Apple]\textsubscript{company}[Watch Series~2]]\textsubscript{product}}}
    \end{itemize}
    \label{ex:rm-examples}
\end{example}


Our annotation guidelines also consider some specific cases. For example, if a sentence contains a full-length company name followed and coreferenced by the company abbreviation, then we label both mentions as individual company mentions, but only a single relation mention, between the full-length company mention and the product, is annotated. The relation between the company acronym and a product is implicitly given by the coreference information. 
Example~\ref{ex:company-full-length-and-acronym} illustrates this issue: the company name \textit{IS International Services LLC}, the trigger \textit{providing} and the product \textit{engineering services} are annotated as relation arguments, while the company's abbreviation \textit{IS} are connected by the \textit{Identity} relation.

\begin{example}
\small
\begin{itemize}
    \item [a.] \textit{\textsf{[IS International Services LLC]\textsubscript{company} ([IS]\textsubscript{company}) is a uniquely qualified business [providing]\textsubscript{trigger} [engineering services]\textsubscript{product}}}
\end{itemize}
\label{ex:company-full-length-and-acronym}
\end{example}

Furthermore, if a sentence contains more than one trigger for the same relation instance, then as many relation mentions are annotated as there are triggers. Example~\ref{ex:multiple-triggers} contains one company, one product, and the three triggers \textit{developer}, \textit{manufacturer} and \textit{vendor}, each of them referring to a different way of how the product is related to the company -- therefore three relation mentions are created.

\begin{example}
\small
\begin{itemize}
	\item [a.] \textit{\textsf{FUJIFILM invested in [Japan Biomedical Co.]\textsubscript{company}, a [developer]\textsubscript{trigger}, [manufacturer]\textsubscript{trigger} and [vendor]\textsubscript{trigger} of [additives for cell culture media]\textsubscript{product}}}.
\end{itemize}
\label{ex:multiple-triggers}
\end{example}

\section{Corpus Statistics}
\label{sec:corpus}

\begin{table}[ht!]
\centering
\begin{tabular}{lrr}
\toprule
 & \textbf{Total} & \textbf{Mean} \\
\midrule
\# Documents & 152 & - \\
\# Sentences & 4001 & 26.3 \\
\# Words & 131929 & 868.0 \\
\midrule
\# Companies & 2191 & 14.4 \\
\# Products & 1717 & 11.3 \\
\# CompanyProvidesProduct & 379 & 2.5 \\
\bottomrule
\end{tabular}
\caption{Corpus Statistics}
\label{tab:dataset-stats}
\end{table}

This section describes the corpus of documents annotated with product mentions, including product parts, technologies, and product classes, using the guidelines described in the previous section. Documents included in the corpus are sampled from the dataset that we described in Section~\ref{sec:bootstrap}  Table~\ref{tab:dataset-stats} lists some statistics of the current state of the corpus. The annotation is being carried out by two trained linguistics students. In cases of disagreement, a third expert annotator is consulted to reach a final decision. The current datasets consists of $152$ documents with more than $131,000$ words. Thus far, $2,908$ entity mentions ($2,191$ organizations, $1,717$ products) have been annotated, and a total of $379$ \emph{CompanyProvidesProduct} relation mentions.

For the corpus annotation we use the markup tool Recon~\cite{Li2012}, which allows annotating n-ary relations among text elements. Recon provides a graphical user interface that enables users to mark arbitrary text spans as entities, to connect entities to create relations, and to assign semantic roles to argument entities. Since the corpus is still in the process of being created, we cannot report any reliable inter-annotator agreement scores at the moment. We will include information about inter-annotator agreement at the entity and relation mention level in the final release. The corpus  and the guidelines will be made available at \url{https://dfki-lt-re-group.bitbucket.io/product-corpus}. We distribute the dataset in an AVRO-based compact binary format, along with the corresponding schema and reader tools.



 \section{Conclusion}
 \label{sec:conclusion}
In this work we presented a fine-grained analysis and annotation schema for mentions of \emph{product} entities and \emph{CompanyProvidesProduct} relations in English web and social media texts. The schema is motivated by linguistic aspects and addresses the needs of recognizing industry- and product-related facts and relations. We presented a semi-automatic annotation process in order to ease the annotation procedure. While we have only annotated a small set of documents so far, the annotation effort to increase the size of the corpus is ongoing. 


\section*{Acknowledgments}
This  research  was  partially  supported  by  the  German  Federal  Ministry  of  Economics  and Energy   (BMWi)   through   the   projects  SD4M (01MD15007B) and SDW (01MD15010A) and by the German Federal Ministry of Education and Research (BMBF) through the project BBDC (01IS14013E).

\section{Bibliographical References}
\label{main:ref}

\bibliographystyle{lrec}
\bibliography{main}


\end{document}